\documentclass[10pt,twocolumn,letterpaper]{article}
\usepackage[pagenumbers]{cvpr}
\usepackage{booktabs,tabularx,array}

\usepackage{colortbl}
\definecolor{CPshade}{gray}{0.94}
\definecolor{CPaccent}{gray}{0}
\newcommand{\CPpanel}[3]{\multicolumn{#1}{@{}l@{}}{\makebox[0pt][l]{\makebox[\textwidth][l]{\textbf{#2}\hfill\textit{#3}}}}}

\DeclareMathSizes{8.4}{8.4}{6}{5}

\definecolor{cvprblue}{rgb}{0.21,0.49,0.74}
\usepackage[pagebackref,breaklinks,colorlinks,allcolors=cvprblue]{hyperref}
\def\confName{CVPR}
\def\confYear{2027}
\title{CrossPanel: Exposing the Answer--Evidence Gap\\in Scientific Figure Understanding}
\author{%
Zhengyang Lu\textsuperscript{1}\thanks{Equal contribution.}\quad
QianhongWen\textsuperscript{2}\footnotemark[1]\quad
Chenpeng Wang\textsuperscript{3}\footnotemark[1]\quad
Qiang Huang\textsuperscript{4}\quad
Ziyang Ling\textsuperscript{1}\\
Huan Li\textsuperscript{1}\quad
Chenziao\textsuperscript{5}\quad
Yibo Qiu\textsuperscript{1}\quad
Xiaojie Cheng\textsuperscript{3}\\[4pt]
{\normalsize
\textsuperscript{1}University of Science and Technology of China\quad
\textsuperscript{2}Sichuan Agricultural University}\\
{\normalsize
\textsuperscript{3}YiXin-AI Lab\quad
\textsuperscript{4}Fudan University\quad
\textsuperscript{5}Peking Union Medical College}\\[2pt]
{\small\texttt{\{leolululu, zaneling, lhlihuan, alexandreqiu\}@mail.ustc.edu.cn}}\\
{\small\texttt{wangchenpeng@yxqiche.com}\quad
\texttt{huangq25@m.fudan.edu.cn}}%
}
\begin{document}
\raggedbottom
\maketitle
\begin{abstract}
A correct answer to a scientific-figure question can conceal an incorrect account of the experiment. Reading evidence across panels requires keeping each observation attached to the object and conditions that give it meaning. We introduce CrossPanel, a benchmark of 2,650 human-reviewed questions that exposes this \emph{answer--evidence gap}. Its tasks combine parallel observations or follow dependent queries, while a response audit separates object selection, experimental conditions, cross-panel binding, and local readout. Across 15 vision--language models on 251 diagnostic questions, 8.42\%--56.64\% of correct answers contain explicit factual errors. The audit distinguishes coherent readings of the wrong target from incorrect readings of the right one. Controlled interventions then test how these failures can be repaired: verified object--condition hints, which withhold the requested measurement, raise Qwen3.5-35B-A3B's target-query success by 22.81 percentage points on 57 questions, while additional image tiles alone yield little improvement. Other models benefit most from combining both forms of assistance. These findings identify experimental attribution as a distinct requirement alongside local perception, and make both measurable beyond final-answer accuracy.
\end{abstract}

\section{Introduction}
\label{sec:intro}
\begin{figure*}[t]
\centering
\includegraphics[width=0.78\textwidth]{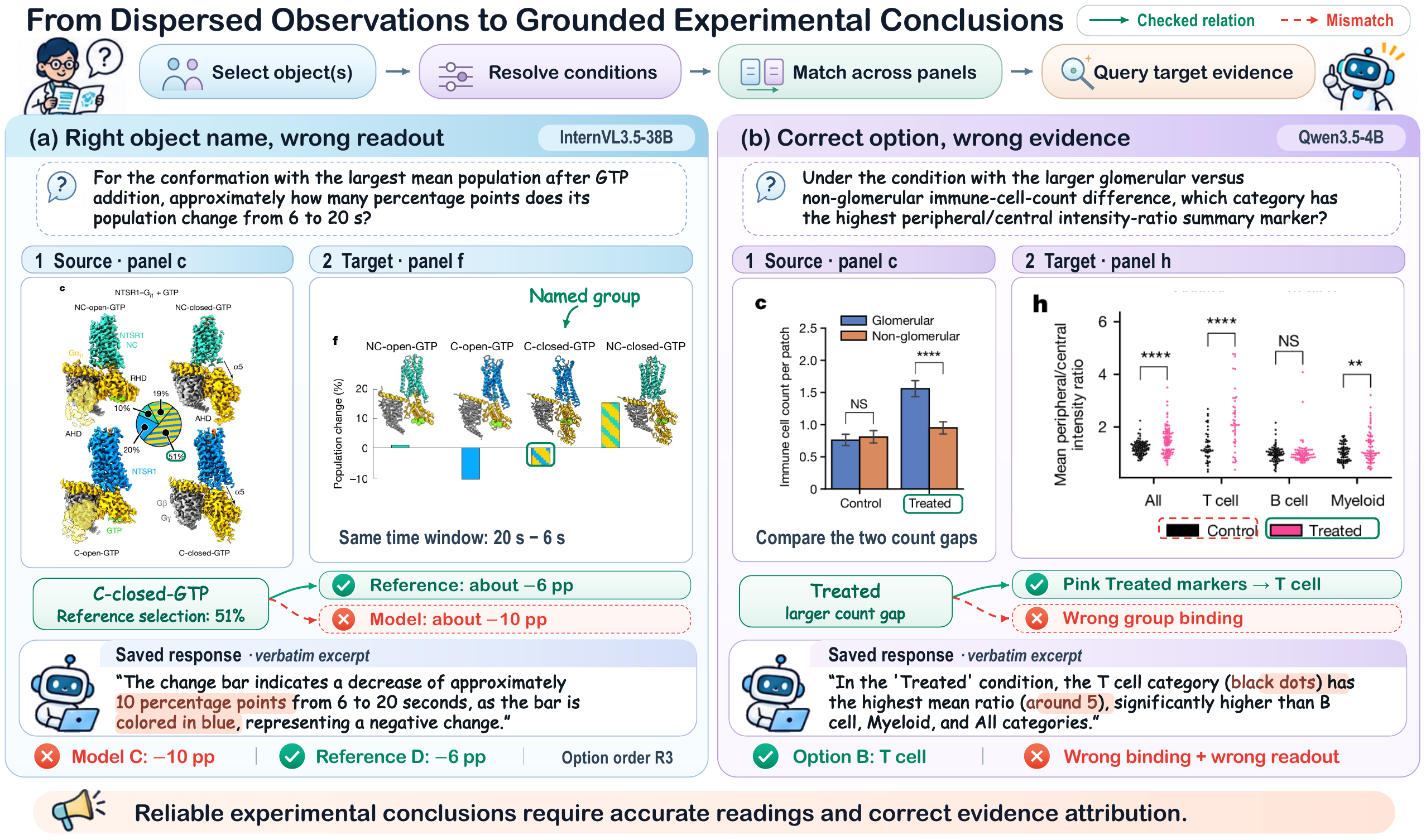}
\caption{\textbf{A correct option can coexist with an incorrect experimental account.}
(a) InternVL3.5-38B names C-closed-GTP correctly but reports a population change of about $-10$ rather than $-6$ percentage points.
(b) Qwen3.5-4B selects the correct T-cell option while assigning it to the wrong condition markers and reporting the wrong readout.
The saved responses reveal errors in experimental attribution and local readout that final-answer scoring alone cannot diagnose.}
\label{fig:overview}
\end{figure*}

A model can select the correct answer to a scientific question while misidentifying the evidence behind it. In Fig.~\ref{fig:overview}, Qwen3.5-4B chooses the correct T-cell option, yet assigns treated samples to control markers and reports the wrong intensity ratio. The answer passes; the experimental account fails. Accuracy alone leaves this contradiction invisible. Shortcut learning likewise shows that successful predictions can conceal an unreliable decision basis~\cite{lapuschkin2019cleverhans,geirhos2020shortcut}.

The difficulty lies in preserving experimental identity across views. A researcher may select a material by its morphology, identify its treatment, and inspect its measured response in another panel. Reading that response accurately is useful only if it belongs to the selected material under the required conditions. Scientific figure understanding therefore couples two capabilities: \emph{evidence attribution}, which determines the relevant object and context, and \emph{local readout}, which extracts the observation.

Scientific and mathematical benchmarks test multimodal knowledge and reasoning across varied visual contexts~\cite{lu2022scienceqa,yue2024mmmu,lu2024mathvista,wang2025mvmath}. SPUR brings fine-grained perception and cross-panel reasoning into evaluation~\cite{ding2026spur}, while Sci-MMR connects evidence and claims through argument graphs~\cite{li2026scimmr}. CrossPanel examines the \emph{answer--evidence gap} within individual responses: when an answer is correct, are the experimental facts supporting it also correct?

CrossPanel pairs 2,650 questions across eight disciplines with a four-dimensional response audit. The audit checks Object, Condition, Binding, and Query separately, allowing a coherent reading of the wrong target to be distinguished from a flawed reading of the right one. Applied to 15 models on 251 questions, it yields 3,765 annotated responses and changes model ordering: DeepSeek V4.1 Flash exceeds GPT-5.4 mini in answer accuracy, but falls behind when answers must be supported by the required facts.

The distinction guides repair. On 57 questions, verified object--condition context improves Qwen3.5-35B-A3B's target-query success by 22.81 points without revealing the measurement; fixed image tiles alone recover little. Qwen3.5-4B and GLM benefit most from combining context and tiles. Diagnosing attribution and readout separately reveals where each form of assistance helps.

Our contributions are:
\begin{itemize}
\item An evidence-centered benchmark of 2,650 scientific-figure questions that tests parallel integration and dependent queries with references linking observations to experimental context.
\item A four-dimensional response audit that exposes the answer--evidence gap and separates attribution errors from local readout errors.
\item A 15-model study with controlled interventions that identifies how experimental context and visual detail repair different failures.
\end{itemize}

\section{Related Work}
\label{sec:related}
\begin{table*}[t]
\centering
\hypersetup{allcolors=black}
\caption{\textbf{What is linked, what is annotated, and what is evaluated.}
Selected neighboring benchmarks are compared by visual context, task relations, reference granularity, and reported assessment.}
\label{tab:positioning}
\begingroup
\color{black}
\definecolor{CPTableTint}{HTML}{EAF2F6}
\setlength{\tabcolsep}{3.5pt}
\renewcommand{\arraystretch}{1.14}
\fontsize{8.4}{9.6}\selectfont
\newcommand{\CPcell}[1]{\begin{tabular}[c]{@{}l@{}}#1\end{tabular}}
\newcommand{\CPstrong}[1]{{\bfseries\CPcell{#1}}}
\newcommand{\CPsmall}[1]{{\fontsize{7}{8}\selectfont\itshape #1}}
\begin{tabular}{@{}p{69pt}p{44pt}p{52pt}p{52pt}p{76pt}p{76pt}p{84pt}@{}}
\toprule
\textbf{Benchmark} & \multicolumn{3}{c}{\textbf{Data \& visual context}} & \multicolumn{3}{c}{\textbf{Task \& evidence assessment}}\\
\cmidrule(lr){2-4}\cmidrule(l){5-7}
 & \textbf{Scale} & \textbf{Domain} & \textbf{Visual input} & \textbf{Relation tested} & \textbf{Reference unit} & \textbf{Reported assessment}\\
\midrule
\multicolumn{7}{@{}l}{\itshape Visual interpretation and scientific QA}\\[1pt]
\CPcell{\textbf{ChartMuseum}~\cite{tang2025chartmuseum}\\\CPsmall{NeurIPS 2025}} & \CPcell{1,162\\\CPsmall{questions}} & \CPcell{Diverse\\topics} & \CPcell{Real-world\\charts} & \CPcell{Visual + textual\\reasoning} & \CPcell{Answers;\\reasoning-type tags} & \CPcell{Answer accuracy;\\visual error analysis}\\[3pt]
\CPcell{\textbf{ChartR}~\cite{chen2026chartr}\\\CPsmall{CVPR 2026}} & \CPcell{200+1,452\\\CPsmall{main + sub-QA}} & \CPcell{20 topic\\domains} & \CPcell{Rendered\\charts} & \CPcell{Dependent\\reasoning steps} & \CPcell{Sub-question\\answer chains} & \CPcell{Step / chain accuracy;\\visual robustness}\\[3pt]
\CPcell{\textbf{MicroVQA}~\cite{burgess2025microvqa}\\\CPsmall{CVPR 2025}} & \CPcell{1,042\\\CPsmall{questions}} & \CPcell{Biology} & \CPcell{Microscopy\\images} & \CPcell{Interpretation;\\research decisions} & \CPcell{Expert-authored\\QA / MCQ} & \CPcell{Answer accuracy;\\expert error analysis}\\[3pt]
\CPcell{\textbf{MMCR}~\cite{tian2025mmcr}\\\CPsmall{ICCV 2025}} & \CPcell{276\\\CPsmall{questions}} & \CPcell{AI / CS} & \CPcell{Scientific\\paper pages} & \CPcell{Cross-source\\reasoning} & \CPcell{MCQ;\\source-type labels} & \CPcell{Circular answer\\accuracy}\\[3pt]
\midrule
\multicolumn{7}{@{}l}{\itshape Grounding and evidence-based reasoning}\\[1pt]
\CPcell{\textbf{MC-Bench}~\cite{xu2025mcbench}\\\CPsmall{ICCV 2025}} & \CPcell{2,000\\\CPsmall{image pairs}} & \CPcell{Mixed visual\\domains} & \CPcell{Paired\\images} & \CPcell{Refer; compare;\\reason} & \CPcell{Instance boxes\\and groups} & \CPcell{Image accuracy;\\instance AP50}\\[3pt]
\CPcell{\textbf{SciVer}~\cite{wang2025sciver}\\\CPsmall{ACL 2025}} & \CPcell{3,000\\\CPsmall{claims}} & \CPcell{CS} & \CPcell{Text, charts\\and tables} & \CPcell{Direct / parallel /\\sequential / analytical} & \CPcell{Evidence IDs:\\text / figure / table} & \CPcell{Entailment accuracy;\\manual error analysis}\\[3pt]
\CPcell{\textbf{SPUR}~\cite{ding2026spur}\\\CPsmall{ACL 2026}} & \CPcell{4,264\\\CPsmall{questions}} & \CPcell{Biomedicine} & \CPcell{Multi-panel\\figures} & \CPcell{Perception; relations;\\scientific reasoning} & \CPcell{MCQ;\\seven task labels} & \CPcell{Task accuracy;\\perception controls}\\[3pt]
\CPcell{\textbf{Sci-MMR}~\cite{li2026scimmr}\\\CPsmall{arXiv 2026}} & \CPcell{235\\\CPsmall{tasks}} & \CPcell{Four scientific\\disciplines} & \CPcell{Figures +\\knowledge} & \CPcell{Multi-hop\\argument graphs} & \CPcell{Evidence / claim\\nodes; regions} & \CPcell{Answer; evidence\\and claim coverage}\\[3pt]
\midrule
\rowcolor{CPTableTint}
\CPcell{\textbf{CrossPanel}\\\CPsmall{Ours}} & \CPcell{2,650\\\CPsmall{questions}} & \CPcell{Science, CS\\\& engineering} & \CPcell{Multi-panel\\figures} & \CPstrong{Parallel integration;\\dependent queries} & \CPstrong{Object, condition,\\binding, query facts} & \CPstrong{Fact-level + joint\\correctness (251 QA)}\\[3pt]
\bottomrule
\end{tabular}
\vspace{2pt}
\begin{minipage}{\textwidth}
\fontsize{7.3}{8.4}\selectfont
Scale follows each task's unit; domain taxonomies differ. ChartR has 200 main and 1,452 sub-questions; 8,260 counts image--question pairs including perturbations.
CrossPanel contains 495 parallel and 2,155 dependent questions; fact-level and joint assessment covers the same 251-question subset for 15 models.
Reported assessment summarizes selected protocols, not an exhaustive inventory of annotations.
\end{minipage}
\endgroup
\end{table*}

\paragraph{From chart answers to scientific evidence.}
Chart question answering established the need to connect visual elements with numerical and logical operations~\cite{kafle2018dvqa,methani2020plotqa,masry2022chartqa}. Recent benchmarks broaden chart diversity and question formats~\cite{wang2024charxiv,masry2025chartqapro,xia2025chartx}, with ChartMuseum emphasizing visual reasoning and SCI-CQA targeting scientific charts~\cite{tang2025chartmuseum,shen2026scicqa}. Scientific interpretation adds the experimental meaning of an observation: MicroVQA links microscopy to research decisions~\cite{burgess2025microvqa}, while SciFIBench tests figure interpretation and probes reasoning faithfulness~\cite{roberts2024scifibench}. SPIQA, MMLongBench-Doc, and MMCR extend evidence gathering across figures, text, and pages~\cite{pramanick2024spiqa,ma2024mmlongbench,tian2025mmcr}. SPUR evaluates perception and reasoning in experimental figures~\cite{ding2026spur}; Sci-MMR reconstructs scientific argument graphs~\cite{li2026scimmr}. CrossPanel audits whether correct answers preserve the identity and conditions of their supporting evidence (Table~\ref{tab:positioning}).

\paragraph{Representing what a figure contains.}
Better chart representations make numerical evidence easier to recover. MatCha and UniChart learn chart-specific structure~\cite{liu2023matcha,masry2023unichart}, DePlot translates plots into tables~\cite{liu2023deplot}, and ChartInstruct extends instruction following across chart tasks~\cite{masry2024chartinstruct}. More diverse synthesis and aligned supervision expand this coverage~\cite{yang2025chartsynthesis,kondic2026chartnet}; Twin-T explicitly separates structural cues from fine details~\cite{bao2026twint}. Scientific image--text resources supply complementary supervision at archive scale~\cite{lozano2025biomedica} and at figure, panel, and region levels~\cite{yuan2026panelpixel}. Marten and M3Grounder further connect document answers with spatial evidence through masks~\cite{wang2025marten,venna2026m3grounder}. Recovering an observation leaves a further question: does it belong to the required experimental context?

\paragraph{Checking the basis of an answer.}
Intermediate verification makes errors visible that outcome scores can miss~\cite{lightman2024verify}, while studies of unfaithful explanations show why a plausible rationale alone is insufficient~\cite{turpin2023unfaithful}. Visual hallucination benchmarks test object existence, attributes, and perceptual confusion~\cite{li2023pope,guan2024hallusionbench,tu2025ode}; HalLoc localizes erroneous tokens and ChartCap evaluates consistency in chart descriptions~\cite{park2025halloc,lim2025chartcap}. For reasoning, MME-CoT evaluates visual and logical steps~\cite{jiang2025mmecot}, ChartR checks dependent sub-questions~\cite{chen2026chartr}, and SciVer grounds scientific claims in evidence~\cite{wang2025sciver}. ProJudge and SVIP assess or supervise individual reasoning steps~\cite{ai2025projudge,gao2025mmcotjudge}; Perceval and recent multimodal CoT analysis target perceptual errors and hallucination-producing steps~\cite{min2026perceptionprm,ma2026mmcothallucination}. CrossPanel audits Object, Condition, Binding, and Query to separate wrong attribution from incorrect local reading.

\paragraph{Finding and inspecting the right evidence.}
Visual programs expose intermediate operations~\cite{gupta2023visprog,suris2023vipergpt}, and ViUniT checks their correctness with visual unit tests~\cite{panagopoulou2025viunit}. Visual Sketchpad and ReFocus let models mark or edit images while reasoning~\cite{hu2024sketchpad,fu2025refocus}; V* and ZoomEye guide search toward relevant details~\cite{wu2024vstar,shen2025zoomeye}. ChartPoint, Chart-FR1, and SketchVL connect chart reasoning to localized visual actions~\cite{xu2025chartpoint,pan2026chartfr1,huang2026sketchvl}, while MC-Bench tests grounding across visual contexts~\cite{xu2025mcbench}. CrossPanel crosses verified context with fixed tiles to distinguish improvements in evidence attribution, local reading, and final answers.

\section{The CrossPanel Benchmark}
\label{sec:method}
\subsection{Experimental context as a task requirement}
\label{sec:task}
CrossPanel tests whether an observation retains its experimental meaning as a query moves across a figure. Each question provides the figure, necessary definitions and panel descriptions in English, and four answer options. Reference records, withheld from the model, link the relevant objects and conditions to evidence locations and observations.

Two task structures capture how that evidence is used. \emph{Parallel evidence integration} combines independently retrievable observations, as when comparing named samples; a \emph{dependent query} uses an earlier finding to select the next target. For example, Fig.~\ref{fig:interventions} selects an arm by its sensory upper quartile at six months, then queries that arm's motor median at twelve months.

\begin{figure*}[t]
\centering
\includegraphics[width=0.82\textwidth]{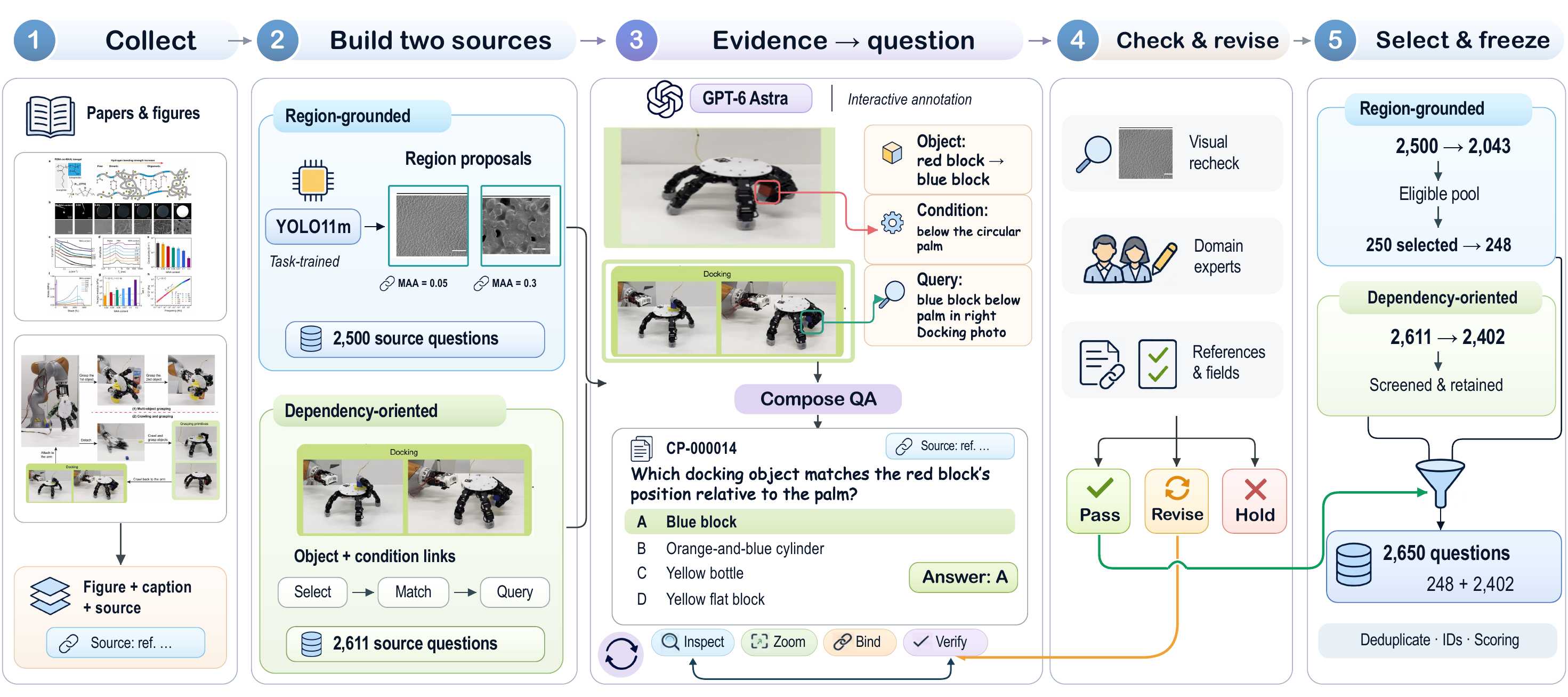}
\caption{\textbf{Complementary coverage of visual evidence and experimental relations.}
Region proposals support local evidence annotation, while object--condition relations connect observations across queries. Human review verifies questions and references against source figures. Selection and screening yield 248 region-grounded and 2,402 dependency-oriented examples.}
\label{fig:construction}

\par\medskip
\includegraphics[width=0.66\textwidth]{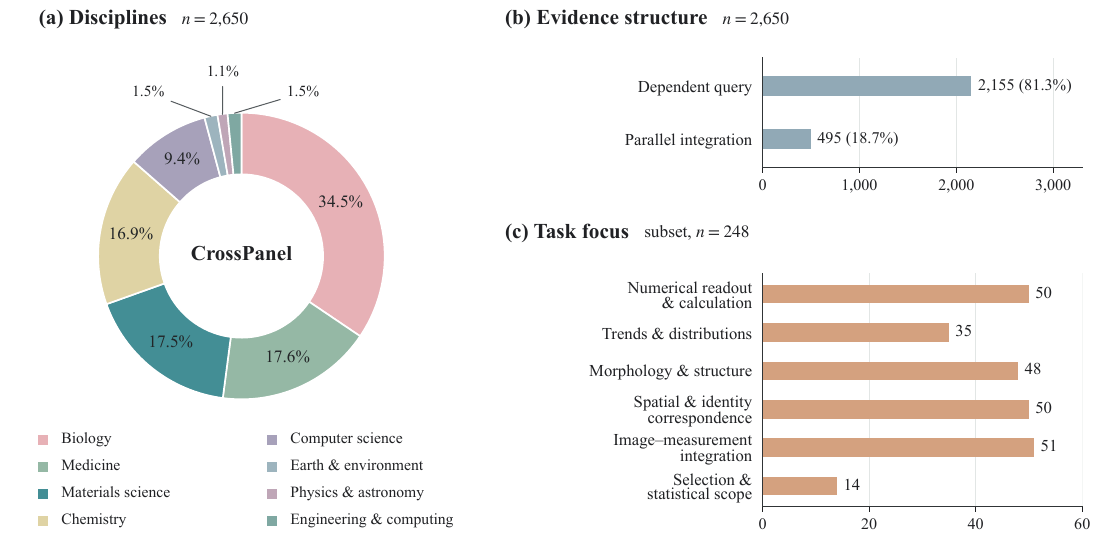}
\caption{\textbf{Coverage and task structure.} Disciplines and evidence structures describe all 2,650 questions; task-focus counts describe the 248-question region-grounded component.}
\label{fig:distribution}
\end{figure*}

\subsection{Complementary evidence coverage}
\label{sec:data}
Two complementary components connect local visual evidence with experimental relations (Fig.~\ref{fig:construction}). The 248-question region-grounded component covers morphology, spatial correspondence, and image--measurement tasks alongside numerical reading and comparison. A task-trained YOLO11m model supports region annotation. The 2,402-question dependency-oriented component extends coverage of object selection and conditions carried across queries.

Human review anchors each retained question and reference to the source figure, following interactive assistance with region inspection and question composition. Together, the components contain 495 parallel and 2,155 dependent tasks across eight disciplines; biology, medicine, materials science, and chemistry account for 86.45\% (Fig.~\ref{fig:distribution}).

\subsection{Diagnostic and controlled evaluation}
The diagnostic suite resolves aggregate scores into evidence-level outcomes. A stratified set of 251 questions spans both sources and task structures, yielding 3,765 annotated responses across 15 models from the main evaluation. A 57-question intervention set tests repairs through context and visual access. Controls appear in Tables~\ref{tab:visual} and~\ref{tab:strategies}.

\section{Evaluation Protocol}
\label{sec:protocol}
\begin{table*}[t]
\centering
\hypersetup{allcolors=black}\color{black}
\caption{\textbf{Task completion, factual support, and answer--evidence reliability.}
All scores are percentages. The main set has 2,650 questions; both diagnostic blocks use the same 251 responses per model.
Macro equally averages the two task categories. Fact support is the proportion graded correct in each dimension; these scores are not a cumulative funnel (Sec.~\ref{sec:metrics}).
Joint requires a correct answer and all four necessary facts supported; $E\!\mid\!A$ is the proportion of answer-correct responses with an explicit factual error.
NR denotes disabled reasoning; GPT-4o uses the 2024-11-20 snapshot. Best and second-best values across all 15 models are \textbf{bold} and \underline{underlined}.}
\label{tab:main}
\begingroup
\small
\setlength{\tabcolsep}{2.4pt}
\renewcommand{\arraystretch}{1.14}
\begin{tabular*}{\textwidth}{@{\extracolsep{\fill}}lrrrr@{\hspace{8pt}}rrrr@{\hspace{8pt}}rrr@{}}
\toprule
 & \multicolumn{4}{c}{\shortstack{\textbf{Task completion}\\$n=2{,}650$}}
 & \multicolumn{4}{c}{\shortstack{\textbf{Fact support}\\$n=251$}}
 & \multicolumn{3}{c}{\shortstack{\textbf{Answer and evidence}\\$n=251$}} \\
\cmidrule(lr){2-5}\cmidrule(lr){6-9}\cmidrule(l){10-12}
\textbf{Model} & \shortstack{Parallel\\495} & \shortstack{Dependent\\2,155} & Macro & Overall & Object & Condition & Binding & Query & Answer & Joint & $E\!\mid\!A\!\downarrow$ \\
\midrule
\rowcolor{CPshade}\multicolumn{12}{c}{\itshape Closed-source models} \\
Gemini 3.5 Flash-Lite & \textbf{85.86} & \textbf{70.53} & \textbf{78.20} & \textbf{73.40} & \underline{85.26} & 90.84 & \textbf{94.42} & \textbf{80.48} & \textbf{75.70} & \textbf{66.14} & \textbf{8.42} \\
GPT-5.4 mini & 81.82 & 64.36 & 73.09 & 67.62 & 84.46 & \textbf{95.22} & \underline{93.63} & \underline{71.71} & 71.31 & \underline{62.55} & \underline{10.61} \\
Grok 4.3\textsuperscript{\scriptsize NR} & 73.33 & 50.95 & 62.14 & 55.13 & 76.89 & 89.64 & 90.84 & 62.15 & 61.35 & 49.00 & 17.53 \\
GPT-4o\textsuperscript{\scriptsize 1120} & 64.85 & 40.42 & 52.63 & 44.98 & 68.13 & 90.84 & 91.24 & 51.79 & 52.99 & 37.45 & 24.81 \\
\addlinespace[3pt]
\rowcolor{CPshade}\multicolumn{12}{c}{\itshape Open-source models} \\
Qwen3.5-35B-A3B & \underline{82.02} & \underline{65.24} & \underline{73.63} & \underline{68.38} & \textbf{86.06} & 92.83 & 92.83 & 68.13 & 68.92 & 56.97 & 17.34 \\
DeepSeek V4.1 Flash\textsuperscript{\scriptsize 1017} & 81.62 & 59.77 & 70.69 & 63.85 & 80.88 & \underline{93.63} & \underline{93.63} & 70.92 & \underline{72.11} & 57.37 & 18.78 \\
Qwen3.5-4B & 70.91 & 51.42 & 61.16 & 55.06 & 79.68 & 87.25 & 92.03 & 63.75 & 60.56 & 48.21 & 19.74 \\
Qwen2.5-VL-72B-Instruct & 65.05 & 44.08 & 54.57 & 48.00 & 65.34 & 81.67 & 86.85 & 50.60 & 53.78 & 35.86 & 20.74 \\
Ministral-3-14B & 63.03 & 40.79 & 51.91 & 44.94 & 72.51 & 88.05 & 88.45 & 54.18 & 52.59 & 36.65 & 27.27 \\
Qwen3-VL-30B-A3B & 62.02 & 39.95 & 50.99 & 44.08 & 68.13 & 83.27 & 85.66 & 52.19 & 53.39 & 34.66 & 34.33 \\
GLM-4.6V-Flash & 62.42 & 39.30 & 50.86 & 43.62 & 74.50 & 80.08 & 85.26 & 47.81 & 49.00 & 32.67 & 30.89 \\
InternVL3.5-38B & 59.39 & 38.79 & 49.09 & 42.64 & 62.55 & 77.69 & 82.07 & 42.63 & 49.80 & 29.08 & 32.80 \\
InternVL3.5-8B & 52.73 & 36.47 & 44.60 & 39.51 & 61.35 & 72.51 & 82.87 & 34.66 & 44.22 & 19.52 & 44.14 \\
Gemma-3-27B-IT & 53.33 & 36.29 & 44.81 & 39.47 & 60.16 & 74.90 & 86.45 & 35.06 & 46.22 & 19.12 & 48.28 \\
Molmo2-8B & 47.27 & 31.60 & 39.44 & 34.53 & 59.36 & 70.52 & 76.10 & 26.69 & 45.02 & 14.34 & 56.64 \\

\bottomrule
\end{tabular*}
\endgroup
\end{table*}

\begin{figure*}[t]
\centering
\includegraphics[width=0.94\textwidth]{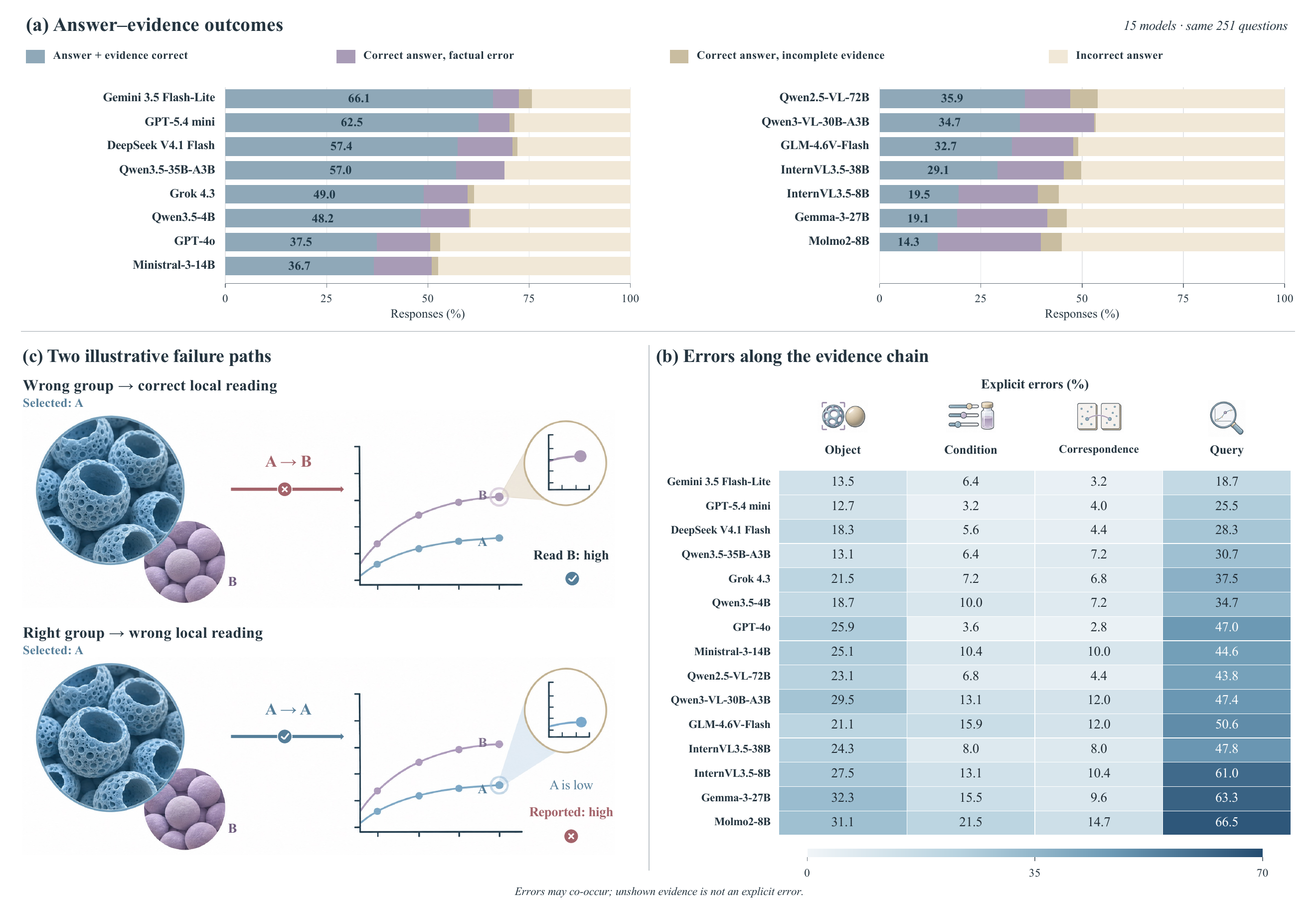}
\caption{\textbf{The answer--evidence gap across 15 models on the same 251 questions.}
(a) Responses are partitioned by answer and fact correctness; incomplete support contains unshown or unresolved required facts without an explicit error.
(b) Schematic examples distinguish wrong attribution with a coherent local readout from an incorrect readout at the correct target. Binding and Query are assessed at the object and location stated in the response.
(c) Each cell is the proportion of all 251 responses with an explicit error in that dimension; errors may co-occur.}
\label{fig:outcomes}
\end{figure*}

\begin{figure*}[t]
\centering
\includegraphics[width=0.78\textwidth]{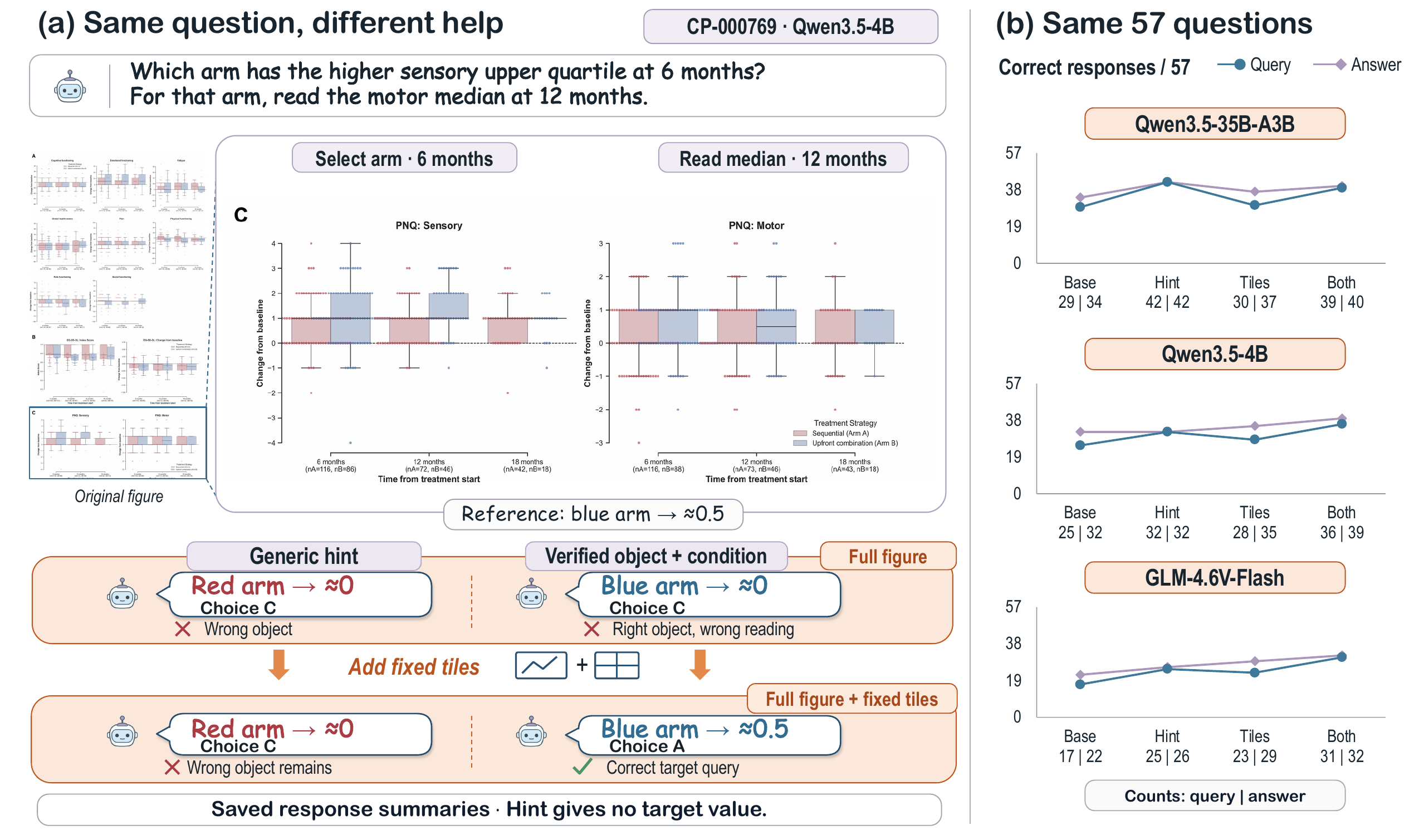}
\caption{\textbf{Context assistance and visual detail repair different failures.}
(a) Saved Qwen3.5-4B responses to the same question illustrate separate object-selection and readout errors.
(b) Correct target queries and answers out of 57 questions. Base: generic hint and full figure; Hint: verified object--condition context and full figure; Tiles: generic hint and full figure plus fixed tiles; Both: verified context plus tiles. Hints provide no target value. Query here requires the question's correct target; lines connect discrete conditions.}
\label{fig:interventions}
\end{figure*}

\begin{table*}[!t]
\centering

\hypersetup{allcolors=black}\color{black}
\caption{\textbf{Images, option content, and order.} A: four-order mean accuracy on 247 questions. All-four and switch rates also use 247; a switch requires distinct nonmissing answer contents. B: long versus end-only options on 241 questions. $\Delta_{\rm img}$ is image minus no-image; $\Delta_{\rm opt}$ is long minus end-only (percentage points). Intervals are 95\% paper-cluster bootstrap intervals.}
\label{tab:visual}
\par\noindent\makebox[\textwidth][c]{\resizebox{0.90\textwidth}{!}{%
\begin{minipage}{\textwidth}
\begingroup
\small
\setlength{\tabcolsep}{4pt}
\renewcommand{\arraystretch}{1.10}
\begin{tabular*}{\textwidth}{@{\extracolsep{\fill}}lrrrrrr@{}}
\toprule
\CPpanel{7}{A. Visual input and order stability}{247 questions $\times$ four orders} \\
 & \multicolumn{4}{c}{\textit{Visual contribution}} & \multicolumn{2}{c}{\textit{Image-present stability}} \\
\cmidrule(lr){2-5}\cmidrule(l){6-7}
\textbf{Model} & Image & No image & $\Delta_{\rm img}$ & 95\% CI & All four $\uparrow$ & Switch $\downarrow$ \\
\midrule
Qwen3.5-35B-A3B & 67.31 & 34.51 & \textbf{+32.79} & [26.62, 38.97] & 48.58 & 33.20 \\
Qwen3.5-4B & 53.54 & 31.07 & \textbf{+22.47} & [16.70, 28.14] & 33.20 & 48.58 \\
Ministral-3-14B & 43.32 & 30.57 & \textbf{+12.75} & [7.99, 18.12] & 21.86 & 59.51 \\
GLM-4.6V-Flash & 38.97 & 30.06 & \textbf{+8.91} & [3.44, 14.38] & 19.43 & 58.30 \\
InternVL3.5-38B & 38.56 & 33.50 & \textbf{+5.06} & [-0.10, 10.32] & 16.60 & 61.13 \\
InternVL3.5-8B & 36.13 & 31.17 & \textbf{+4.96} & [-0.10, 10.12] & 14.57 & 67.21 \\

\bottomrule
\end{tabular*}
\par\vspace{5pt}
\begin{tabular*}{\textwidth}{@{\extracolsep{\fill}}lrrrrrr@{}}
\toprule
\CPpanel{7}{B. Information carried by the options}{241 questions $\times$ four orders} \\
 & \multicolumn{3}{c}{\textit{With image}} & \multicolumn{3}{c}{\textit{Without image}} \\
\cmidrule(lr){2-4}\cmidrule(l){5-7}
\textbf{Model} & Long & End-only & $\Delta_{\rm opt}$ & Long & End-only & $\Delta_{\rm opt}$ \\
\midrule
Qwen3.5-35B-A3B & 75.62 & 67.53 & \textbf{+8.09} & 41.80 & 34.44 & \textbf{+7.37} \\
Qwen3.5-4B & 64.21 & 53.42 & \textbf{+10.79} & 36.00 & 30.71 & \textbf{+5.29} \\
GLM-4.6V-Flash & 51.56 & 38.69 & \textbf{+12.86} & 34.44 & 29.98 & \textbf{+4.46} \\

\bottomrule
\end{tabular*}
\endgroup

\end{minipage}}}

\par\medskip

\hypersetup{allcolors=black}\color{black}
\caption{\textbf{Response strategies and tools.} A: 247 questions, one preselected order, full figures, greedy decoding, and a 2,048-token cap. Acc./No choice are percentages; Tok. is mean output length. Bold marks each model's best accuracy. B: 57 questions and a four-call ceiling; actual costs differ. Target Q requires the requested observation; Joint requires a correct answer and four supported facts. ReFocus~\cite{fu2025refocus} and the SPUR-style VoT prompt~\cite{ding2026spur} are adaptations.}
\label{tab:strategies}
\par\noindent\makebox[\textwidth][c]{\resizebox{0.90\textwidth}{!}{%
\begin{minipage}{\textwidth}
\begingroup
\small
\setlength{\tabcolsep}{3.4pt}
\renewcommand{\arraystretch}{1.10}
\begin{tabular*}{\textwidth}{@{\extracolsep{\fill}}lrrr@{\hspace{12pt}}rrr@{\hspace{12pt}}rrr@{}}
\toprule
\CPpanel{10}{A. Response strategies}{247 questions; one call per question} \\
 & \multicolumn{3}{c}{\textbf{Qwen3.5-4B}} & \multicolumn{3}{c}{\textbf{Qwen3.5-35B-A3B}} & \multicolumn{3}{c}{\textbf{GLM-4.6V-Flash}} \\
\cmidrule(lr){2-4}\cmidrule(lr){5-7}\cmidrule(l){8-10}
\textbf{Instruction} & Acc. $\uparrow$ & No choice $\downarrow$ & Tok. & Acc. $\uparrow$ & No choice $\downarrow$ & Tok. & Acc. $\uparrow$ & No choice $\downarrow$ & Tok. \\
\midrule
Brief explanation & 53.04 & 5.26 & 454 & \textbf{67.61} & 6.48 & 557 & \textbf{38.06} & 0.00 & 102 \\
Answer only & 36.44 & 6.88 & 227 & 48.18 & 2.43 & 101 & 34.82 & 0.00 & 4 \\
CoT & \textbf{55.06} & 12.96 & 656 & 63.16 & 10.93 & 608 & 19.84 & 55.06 & 205 \\
Four fields & 54.66 & 3.64 & 217 & 50.61 & 1.21 & 111 & 5.26 & 84.62 & 22 \\
VoT prompt & 53.44 & 15.79 & 711 & 65.59 & 12.55 & 689 & 8.91 & 80.57 & 203 \\

\bottomrule
\end{tabular*}
\par\vspace{5pt}
\begin{tabular*}{\textwidth}{@{\extracolsep{\fill}}llrrrrrr@{}}
\toprule
\CPpanel{8}{B. Text review versus visual tools}{57 questions; four-call ceiling} \\
\textbf{Model} & \textbf{Method} & Acc. & Target Q & Joint & Calls/q & In. tok./q & Out. tok./q \\
\midrule
Qwen3.5-4B & Text review & 47.37 & 36.84 & 33.33 & 4.00 & 9,728 & 1,191 \\
 & ReFocus (adapt.) & 49.12 & 38.60 & 33.33 & 1.98 & 8,009 & 1,589 \\
\addlinespace[3pt]
GLM-4.6V-Flash & Text review & 40.35 & 26.32 & 21.05 & 4.00 & 9,348 & 409 \\
 & ReFocus (adapt.) & 38.60 & 29.82 & 21.05 & 2.26 & 9,397 & 893 \\

\bottomrule
\end{tabular*}
\endgroup

\end{minipage}}}

\end{table*}

\subsection{Answer and fact-level correctness}
\label{sec:metrics}
Evaluation separates task completion from the evidence supporting it. Answer correctness $A_i$ compares the selected \emph{content} with the reference under the displayed option order. A gold-blind parser extracts final selections, counting missing or conflicting choices as incorrect. Overall accuracy is $N^{-1}\sum_i A_i$; Macro equally averages the two task-structure accuracies.

The response audit follows four links between an answer and its evidence. \emph{Object} ($O$) identifies the selected entity, candidate set, or intermediate anchor; \emph{Condition} ($C$) specifies the required context and comparison rule, such as treatment, time, or mean versus maximum. \emph{Binding} ($B$), the cross-panel correspondence, links the selected object to its evidence, while \emph{Query} ($Q$) evaluates the local reading, comparison, or calculation at the stated location.

Separating attribution from readout requires assessing Binding and Query at the response's selected object and location: a model can choose the wrong species yet correctly follow and read that species' plot. Each dimension is labeled supported (\texttt{c}), explicit error (\texttt{e}), unshown (\texttt{o}), or unresolved (\texttt{u}). Distinguishing contradictions with the figure from omitted details keeps factual errors separate from incomplete explanations.

For labels $g_{if}$, joint correctness and error among correct answers are
\begin{align}
J_i &= A_i\prod_{f\in\{O,C,B,Q\}}\mathbf{1}[g_{if}=\texttt{c}],\\
E\!\mid\!A &= \frac{\sum_i A_i\,\mathbf{1}[\exists f:g_{if}=\texttt{e}]}{\sum_i A_i}.
\end{align}
Joint accuracy averages $J_i$ over the 251 diagnostic questions; per-dimension support and error rates use the same denominator. Human-reviewed reference records and source figures anchor the assessment, with GPT-6 Astra serving as the LLM judge through Codex. Building on model-based evaluation~\cite{zheng2023mtbench,liu2023geval}, the rubric specifies factual criteria for each dimension, following the emphasis on chart-specific and criterion-level judgments~\cite{laskar2025chartjudges,xiong2026multicrit}.

\subsection{Controls on input and assistance}
\label{sec:controls}
Image-present and image-absent conditions use the same questions and four cyclic option orders~\cite{tian2025mmcr}, with accuracy averaged over orders within each question. \emph{All-four} measures consistent correctness, while a \emph{switch} records distinct nonmissing answer contents across orders; missing selections are tracked separately. Long-option controls restore descriptive content omitted from the main evaluation's end-only options.

A $2\times2$ intervention tests whether restoring context or visual detail repairs the requested observation. On 57 questions, a verified object--condition hint or a generic reminder accompanies either the full figure or the full figure plus fixed overlapping quadrants. Verified hints withhold the target observation; tiles increase visual input. Here, \emph{target-query} success requires the correct observation at the question's verified target, combining target selection with local reading.

Paired differences use 95\% intervals obtained by resampling source-paper groups with all their observations and option orders; the intervention set comprises 57 groups. Response strategies share a 2,048-token cap, and tool comparisons share a four-call ceiling with actual calls and tokens reported.

\section{Experiments}
\label{sec:experiments}
\subsection{Evaluation setup and task performance}
We evaluate 15 models with a common task interface: the full figure, English question context, end-only options, and a request for a brief explanation and explicit choice. Ten models run locally with greedy decoding and a 2,048-token cap; five use APIs. External image limits are a 2,048-pixel long edge and 2,097,152 total pixels. API settings include disabled reasoning for Grok 4.3 and the 2024-11-20 GPT-4o snapshot. The supplement reports a previously evaluated Gemini 3 Flash Preview endpoint outside this cohort.

Dependent queries have lower pooled accuracy for every model (Table~\ref{tab:main}). Within the dependency-oriented source, parallel accuracy remains higher by 4.20--13.11 points for 14 models, with Molmo2 nearly tied. Gemini 3.5 Flash-Lite leads this cohort overall at 73.40\%, followed by Qwen3.5-35B-A3B at 68.38\% and GPT-5.4 mini at 67.62\%.

\subsection{The answer--evidence gap}
\label{sec:answerfacts}
Correct selections frequently survive contradictions in their stated evidence. On the same 251 diagnostic responses per model, 8.42\%--56.64\% of correct answers contain explicit factual errors (Fig.~\ref{fig:outcomes}). Even Gemini 3.5 Flash-Lite, which leads both diagnostic scores, drops from 75.70\% answer accuracy to 66.14\% joint correctness: of 190 correct answers, 16 contain explicit errors and eight lack complete support.

Requiring factual support reverses model ordering. DeepSeek V4.1 Flash exceeds GPT-5.4 mini in answer accuracy (72.11\% versus 71.31\%), but falls behind in joint correctness (57.37\% versus 62.55\%). Evidence support thus changes which model the ranking favors.

\subsection{Separating attribution from readout}
Local reading is a major source of failure: Query has the highest explicit-error rate of the four dimensions for every model, ranging from 18.73\% to 66.53\% (Fig.~\ref{fig:outcomes}).

A coherent readout can follow the wrong object. In a human--mouse morphology question, Qwen3.5-35B selects the human group at about 40 features and correctly follows its plot to the longest Avg.\ Straightness bar, although the task requires the mouse group. Conversely, a correct target can lead to a flawed readout: the model selects the lowest-glass-transition material, $\mathrm{phen}_{1.33}|\mathrm{ZIF}$-12, and its heating PXRD sequence, yet places the loss of sharp peaks at 150--200$^\circ$C instead of 200--230$^\circ$C. Correcting attribution addresses the first failure; the second requires better reading at an already identified target.

\subsection{Context and visual access as repairs}
\label{sec:interventions}
Restoring experimental context substantially improves Qwen3.5-35B's target-query success (Fig.~\ref{fig:interventions}). On 57 questions, verified object--condition hints raise success from 29 to 42 cases, a gain of 22.81 points with a paired 95\% interval of [10.53, 36.84]. Answer accuracy also rises by 14.04 points [3.51, 24.56]. Because the hints withhold the requested measurement, the recovered queries show that supplying the right context enables readouts the model failed to produce from the original input.

Extra views alone do little for this model: fixed tiles change target-query success from 29 to 30, and adding them to verified context gives 39 successes versus 42 with context alone (difference interval [$-15.79$, 5.26] points). Context and visual detail complement each other more strongly for Qwen3.5-4B and GLM, whose highest query counts occur with both. For Qwen3.5-4B, context alone improves queries from 25 to 32 while answers remain at 32; combining context and tiles raises the counts to 36 and 39. Query repair can thus precede improvement in the final choice.

The case in Fig.~\ref{fig:interventions} makes this complementarity concrete. Verified context directs Qwen3.5-4B to the required blue arm, but its twelve-month motor median is still misread as approximately zero. Tiles alone leave the wrong arm selected; combining both yields the correct arm and a readout of approximately 0.5.

\subsection{Input controls and general-purpose assistance}
Stable aggregate scores can hide unstable judgments. Qwen3.5-35B scores 66.80--67.61\% across option orders, yet switches answer content on 82/247 questions (Table~\ref{tab:visual}). Images improve mean accuracy by 4.96--32.79 points, with the two InternVL intervals including zero, while longer options help both with and without images.

General-purpose assistance yields inconsistent gains (Table~\ref{tab:strategies}). Qwen3.5-4B gains 2.02 points from CoT~\cite{wei2022cot} [$-3.64$, 7.69], while GLM's large prompt losses accompany many unextractable choices. On 57 questions, the ReFocus adaptation changes correct answers by $+1$ for Qwen3.5-4B and $-1$ for GLM relative to text review, with both paired intervals including zero. Joint correctness remains unchanged: these shifts in final choices do not improve the evidence supporting them.

\section{Discussion and Conclusion}
\label{sec:conclusion}
CrossPanel exposes a gap between selecting the correct answer and correctly accounting for the scientific evidence behind it. Auditing Object, Condition, Binding, and Query within the same response separates errors in experimental attribution from failures of local reading. Across 15 models, this distinction reveals factual errors hidden by answer accuracy and changes model ordering, making evidence support an essential part of evaluating scientific figure understanding.

The interventions turn this diagnosis into a design direction. Verified context restores observations that extra image tiles alone scarcely recover for Qwen3.5-35B, while Qwen3.5-4B and GLM benefit most from combining both. These complementary gains motivate systems that preserve object--condition links while checking the observations reached through them. CrossPanel provides a way to measure that progress: whether a model's improved answer is accompanied by a stronger experimental basis. The goal is to reach the right conclusion \emph{with the right experimental evidence}.

\clearpage
\section*{LLM Usage Statement}
Interactive model assistance supported question formulation and evidence annotation, followed by human review of dataset questions and references. GPT-6 Astra, accessed through Codex, served as the LLM judge for model-response assessment and adjudication. Codex also assisted with reference inspection, analysis of saved records, literature verification, and manuscript drafting and revision. GPT Image generated icons used in the five figures. Experimental examples and numerical results derive from source figures and saved responses; illustrative schematics are identified separately. The authors are responsible for the final references, annotations, results, and text.

{\small
\bibliographystyle{ieeenat_fullname}
\bibliography{main}

\begin{thebibliography}{62}
\providecommand{\natexlab}[1]{#1}
\providecommand{\url}[1]{\texttt{#1}}
\expandafter\ifx\csname urlstyle\endcsname\relax
  \providecommand{\doi}[1]{doi: #1}\else
  \providecommand{\doi}{doi: \begingroup \urlstyle{rm}\Url}\fi

\bibitem[Ai et~al.(2025)Ai, Zhou, Xu, Li, Zhang, Li, Sun, Feng, Huang, Wang, and Zhang]{ai2025projudge}
Jiaxin Ai, Pengfei Zhou, Zhaopan Xu, Ming Li, Fanrui Zhang, Zizhen Li, Jianwen Sun, Yukang Feng, Baojin Huang, Zhongyuan Wang, and Kaipeng Zhang.
\newblock {ProJudge: A Multi-Modal Multi-Discipline Benchmark and Instruction-Tuning Dataset for MLLM-based Process Judges}.
\newblock In \emph{Proceedings of the IEEE/CVF International Conference on Computer Vision (ICCV)}, pages 4681--4690, 2025.

\bibitem[Bao et~al.(2026)Bao, Cheng, Du, Xia, He, Lang, and Liu]{bao2026twint}
Jiahua Bao, Siyao Cheng, Jiaxing Du, Qingtao Xia, Changjiang He, Zeming Lang, and Jie Liu.
\newblock {Twin-T \& TwintVQA: A Reliable Structure-Detail Separating VLM and a Comprehensive Benchmark for Chart and Table Tasks}.
\newblock In \emph{Proceedings of the IEEE/CVF Conference on Computer Vision and Pattern Recognition (CVPR)}, pages 4850--4859, 2026.

\bibitem[Burgess et~al.(2025)Burgess, Nirschl, Bravo-S\'anchez, Lozano, Gupte, Galaz-Montoya, Zhang, Su, Bhowmik, Coman, Hasan, Johannesson, Leineweber, Nair, Yarlagadda, Zuraski, Chiu, Cohen, Hansen, Leonetti, Liu, Lundberg, and Yeung-Levy]{burgess2025microvqa}
James Burgess, Jeffrey~J Nirschl, Laura Bravo-S\'anchez, Alejandro Lozano, Sanket~Rajan Gupte, Jesus~G. Galaz-Montoya, Yuhui Zhang, Yuchang Su, Disha Bhowmik, Zachary Coman, Sarina~M Hasan, Alexandra Johannesson, William~D. Leineweber, Malvika~G Nair, Ridhi Yarlagadda, Connor Zuraski, Wah Chiu, Sarah Cohen, Jan~N. Hansen, Manuel~D Leonetti, Chad Liu, Emma Lundberg, and Serena Yeung-Levy.
\newblock {{MicroVQA}: A Multimodal Reasoning Benchmark for Microscopy-Based Scientific Research}.
\newblock In \emph{Proceedings of the IEEE/CVF Conference on Computer Vision and Pattern Recognition (CVPR)}, pages 19552--19564, 2025.

\bibitem[Chen et~al.(2026)Chen, Luo, Liu, Yang, Zhang, and Zhang]{chen2026chartr}
Xiaojun Chen, Sixiao Luo, Ziqi Liu, Min Yang, Qin Zhang, and Liang-Jie Zhang.
\newblock {{ChartR}: Evaluating Reasoning Accuracy and Robustness in Chart Question Answering}.
\newblock In \emph{Proceedings of the IEEE/CVF Conference on Computer Vision and Pattern Recognition (CVPR)}, pages 41193--41202, 2026.

\bibitem[Ding et~al.(2026)Ding, Tang, E, Ji, Liu, Tian, Sun, Sun, Xu, Liu, Gao, Xi, Jiang, Zhao, Li, Li, Liu, Yang, Chen, and Lin]{ding2026spur}
Junpeng Ding, Zichen Tang, Haihong E, Mengyuan Ji, Yang Liu, Haolin Tian, Haiyang Sun, Pengqi Sun, Yang Xu, Yichen Liu, Haocheng Gao, Zijie Xi, Ruomeng Jiang, Peizhi Zhao, Rongjin Li, Yuanze Li, Jiacheng Liu, Zhongjun Yang, Jintong Chen, and Siying Lin.
\newblock {Decoding Scientific Experimental Images: The {SPUR} Benchmark for Perception, Understanding, and Reasoning}.
\newblock In \emph{Proceedings of the 64th Annual Meeting of the {A}ssociation for {C}omputational {L}inguistics (Volume 1: Long Papers)}, pages 12859--12882, 2026.

\bibitem[Fu et~al.(2025)Fu, Liu, Yang, Corring, Lu, Yang, Roth, Florencio, and Zhang]{fu2025refocus}
Xingyu Fu, Minqian Liu, Zhengyuan Yang, John~Richard Corring, Yijuan Lu, Jianwei Yang, Dan Roth, Dinei Florencio, and Cha Zhang.
\newblock {{R}e{F}ocus: Visual Editing as a Chain of Thought for Structured Image Understanding}.
\newblock In \emph{Proceedings of the 42nd International Conference on Machine Learning}, pages 17783--17805, 2025.

\bibitem[Gao et~al.(2025)Gao, Liu, Yue, Wu, Chen, Li, Tang, Wu, Chua, and Zhuang]{gao2025mmcotjudge}
Minghe Gao, Xuqi Liu, Zhongqi Yue, Yang Wu, Shuang Chen, Juncheng Li, Siliang Tang, Fei Wu, Tat-Seng Chua, and Yueting Zhuang.
\newblock {Benchmarking Multimodal CoT Reward Model Stepwise by Visual Program}.
\newblock In \emph{Proceedings of the IEEE/CVF International Conference on Computer Vision (ICCV)}, pages 1718--1728, 2025.

\bibitem[Geirhos et~al.(2020)Geirhos, Jacobsen, Michaelis, Zemel, Brendel, Bethge, and Wichmann]{geirhos2020shortcut}
Robert Geirhos, J{\"{o}}rn-Henrik Jacobsen, Claudio Michaelis, Richard Zemel, Wieland Brendel, Matthias Bethge, and Felix~A. Wichmann.
\newblock {Shortcut learning in deep neural networks}.
\newblock \emph{Nature Machine Intelligence}, 2\penalty0 (11):\penalty0 665--673, 2020.

\bibitem[Guan et~al.(2024)Guan, Liu, Wu, Xian, Li, Liu, Wang, Chen, Huang, Yacoob, Manocha, and Zhou]{guan2024hallusionbench}
Tianrui Guan, Fuxiao Liu, Xiyang Wu, Ruiqi Xian, Zongxia Li, Xiaoyu Liu, Xijun Wang, Lichang Chen, Furong Huang, Yaser Yacoob, Dinesh Manocha, and Tianyi Zhou.
\newblock {HallusionBench: An Advanced Diagnostic Suite for Entangled Language Hallucination and Visual Illusion in Large Vision-Language Models}.
\newblock In \emph{Proceedings of the IEEE/CVF Conference on Computer Vision and Pattern Recognition (CVPR)}, pages 14375--14385, 2024.

\bibitem[Gupta and Kembhavi(2023)]{gupta2023visprog}
Tanmay Gupta and Aniruddha Kembhavi.
\newblock {Visual Programming: Compositional Visual Reasoning Without Training}.
\newblock In \emph{Proceedings of the IEEE/CVF Conference on Computer Vision and Pattern Recognition (CVPR)}, pages 14953--14962, 2023.

\bibitem[Hu et~al.(2024)Hu, Shi, Fu, Roth, Ostendorf, Zettlemoyer, Smith, and Krishna]{hu2024sketchpad}
Yushi Hu, Weijia Shi, Xingyu Fu, Dan Roth, Mari Ostendorf, Luke Zettlemoyer, Noah Smith, and Ranjay Krishna.
\newblock {Visual Sketchpad: Sketching as a Visual Chain of Thought for Multimodal Language Models}.
\newblock In \emph{Advances in Neural Information Processing Systems}, pages 139348--139379, 2024.

\bibitem[Huang et~al.(2026)Huang, Zhang, Li, Wu, and Liu]{huang2026sketchvl}
Muye Huang, Lingling Zhang, Yifei Li, Yaqiang Wu, and Jun Liu.
\newblock {SketchVL: Policy Optimization via Fine-Grained Credit Assignment for Chart Understanding and More}.
\newblock In \emph{Proceedings of the IEEE/CVF Conference on Computer Vision and Pattern Recognition (CVPR)}, pages 4738--4748, 2026.

\bibitem[Jiang et~al.(2025)Jiang, Zhang, Guo, Li, Qi, Chen, Wang, Jin, Guo, Yan, Zhang, Fu, Gao, and Li]{jiang2025mmecot}
Dongzhi Jiang, Renrui Zhang, Ziyu Guo, Yanwei Li, Yu Qi, Xinyan Chen, Liuhui Wang, Jianhan Jin, Claire Guo, Shen Yan, Bo Zhang, Chaoyou Fu, Peng Gao, and Hongsheng Li.
\newblock {{MME}-{C}o{T}: Benchmarking Chain-of-Thought in Large Multimodal Models for Reasoning Quality, Robustness, and Efficiency}.
\newblock In \emph{Proceedings of the 42nd International Conference on Machine Learning}, pages 27793--27830, 2025.

\bibitem[Kafle et~al.(2018)Kafle, Price, Cohen, and Kanan]{kafle2018dvqa}
Kushal Kafle, Brian Price, Scott Cohen, and Christopher Kanan.
\newblock {DVQA: Understanding Data Visualizations via Question Answering}.
\newblock In \emph{Proceedings of the IEEE Conference on Computer Vision and Pattern Recognition (CVPR)}, pages 5648--5656, 2018.

\bibitem[Kondic et~al.(2026)Kondic, Li, Joshi, Sanchez, Wiesel, Abedin, Alfassy, Schwartz, Caraballo, Cinar, Scheidegger, Ross, Weidele, Hua, Arutyunova, Herzig, Wang, Yu, Zhao, Jiang, Liu, Lin, Oliva, and Feris]{kondic2026chartnet}
Jovana Kondic, Pengyuan Li, Dhiraj Joshi, Isaac Sanchez, Ben Wiesel, Shafiq Abedin, Amit Alfassy, Eli Schwartz, Daniel Caraballo, Yagmur~Gizem Cinar, Florian Scheidegger, Steven~I. Ross, Daniel Karl~I. Weidele, Hang Hua, Ekaterina Arutyunova, Roei Herzig, Zihan Wang, Xinyue Yu, Yunfei Zhao, Sicong Jiang, Minghao Liu, Qunshu Lin, Aude Oliva, and Rogerio Feris.
\newblock {ChartNet: A Million-Scale, High-Quality Multimodal Dataset for Robust Chart Understanding}.
\newblock In \emph{Proceedings of the IEEE/CVF Conference on Computer Vision and Pattern Recognition (CVPR)}, pages 15922--15932, 2026.

\bibitem[Lapuschkin et~al.(2019)Lapuschkin, W{\"{a}}ldchen, Binder, Montavon, Samek, and M{\"{u}}ller]{lapuschkin2019cleverhans}
Sebastian Lapuschkin, Stephan W{\"{a}}ldchen, Alexander Binder, Gr{\'{e}}goire Montavon, Wojciech Samek, and Klaus-Robert M{\"{u}}ller.
\newblock {Unmasking Clever Hans predictors and assessing what machines really learn}.
\newblock \emph{Nature Communications}, 10\penalty0 (1):\penalty0 1096, 2019.

\bibitem[Laskar et~al.(2025)Laskar, Islam, Mahbub, Masry, Rahman, Bhuiyan, Nayeem, Joty, Hoque, and Huang]{laskar2025chartjudges}
Md~Tahmid~Rahman Laskar, Mohammed~Saidul Islam, Ridwan Mahbub, Ahmed Masry, Mizanur Rahman, Amran Bhuiyan, Mir~Tafseer Nayeem, Shafiq Joty, Enamul Hoque, and Jimmy~Xiangji Huang.
\newblock {Judging the Judges: Can Large Vision-Language Models Fairly Evaluate Chart Comprehension and Reasoning?}
\newblock In \emph{Proceedings of the 63rd Annual Meeting of the {A}ssociation for {C}omputational {L}inguistics (Volume 6: Industry Track)}, pages 1203--1216, 2025.

\bibitem[Li et~al.(2026)Li, Yang, Xi, Chen, Zhou, Jin, Nan, Zhang, Wang, Li, Zhu, Deng, Wang, Zheng, Zhang, Bai, Ma, and Gui]{li2026scimmr}
Jiaqiang Li, Yajie Yang, Zhiheng Xi, Jiadong Chen, Enyu Zhou, Senjie Jin, Yang Nan, Jiazheng Zhang, Han Wang, Yanxin Li, Dingwei Zhu, Bicheng Deng, Yuhui Wang, Xiang Zheng, Qi Zhang, Lei Bai, Xingjun Ma, and Tao Gui.
\newblock {{Sci-MMR}: Benchmarking Multi-Step Evidence-Grounded Scientific Reasoning in Multimodal Agents}, 2026.
\newblock arXiv preprint arXiv:2609.11243.

\bibitem[Li et~al.(2023)Li, Du, Zhou, Wang, Zhao, and Wen]{li2023pope}
Yifan Li, Yifan Du, Kun Zhou, Jinpeng Wang, Wayne~Xin Zhao, and Ji-Rong Wen.
\newblock {Evaluating Object Hallucination in Large Vision-Language Models}.
\newblock In \emph{Proceedings of the 2023 Conference on Empirical Methods in Natural Language Processing}, pages 292--305, 2023.

\bibitem[Lightman et~al.(2024)Lightman, Kosaraju, Burda, Edwards, Baker, Lee, Leike, Schulman, Sutskever, and Cobbe]{lightman2024verify}
Hunter Lightman, Vineet Kosaraju, Yuri Burda, Harrison Edwards, Bowen Baker, Teddy Lee, Jan Leike, John Schulman, Ilya Sutskever, and Karl Cobbe.
\newblock {Let's Verify Step by Step}.
\newblock In \emph{International Conference on Learning Representations}, pages 39578--39601, 2024.

\bibitem[Lim et~al.(2025)Lim, Ahn, and Kim]{lim2025chartcap}
Junyoung Lim, Jaewoo Ahn, and Gunhee Kim.
\newblock {ChartCap: Mitigating Hallucination of Dense Chart Captioning}.
\newblock In \emph{Proceedings of the IEEE/CVF International Conference on Computer Vision (ICCV)}, pages 13171--13182, 2025.

\bibitem[Liu et~al.(2023{\natexlab{a}})Liu, Eisenschlos, Piccinno, Krichene, Pang, Lee, Joshi, Chen, Collier, and Altun]{liu2023deplot}
Fangyu Liu, Julian Eisenschlos, Francesco Piccinno, Syrine Krichene, Chenxi Pang, Kenton Lee, Mandar Joshi, Wenhu Chen, Nigel Collier, and Yasemin Altun.
\newblock {{D}e{P}lot: One-shot visual language reasoning by plot-to-table translation}.
\newblock In \emph{Findings of the {A}ssociation for {C}omputational {L}inguistics: {ACL} 2023}, pages 10381--10399, 2023{\natexlab{a}}.

\bibitem[Liu et~al.(2023{\natexlab{b}})Liu, Piccinno, Krichene, Pang, Lee, Joshi, Altun, Collier, and Eisenschlos]{liu2023matcha}
Fangyu Liu, Francesco Piccinno, Syrine Krichene, Chenxi Pang, Kenton Lee, Mandar Joshi, Yasemin Altun, Nigel Collier, and Julian Eisenschlos.
\newblock {{M}at{C}ha: Enhancing Visual Language Pretraining with Math Reasoning and Chart Derendering}.
\newblock In \emph{Proceedings of the 61st Annual Meeting of the {A}ssociation for {C}omputational {L}inguistics (Volume 1: Long Papers)}, pages 12756--12770, 2023{\natexlab{b}}.

\bibitem[Liu et~al.(2023{\natexlab{c}})Liu, Iter, Xu, Wang, Xu, and Zhu]{liu2023geval}
Yang Liu, Dan Iter, Yichong Xu, Shuohang Wang, Ruochen Xu, and Chenguang Zhu.
\newblock {{G}-Eval: {NLG} Evaluation using {G}pt-4 with Better Human Alignment}.
\newblock In \emph{Proceedings of the 2023 Conference on Empirical Methods in Natural Language Processing}, pages 2511--2522, 2023{\natexlab{c}}.

\bibitem[Lozano et~al.(2025)Lozano, Sun, Burgess, Chen, Nirschl, Gu, Lopez, Aklilu, Rau, Katzer, Zhang, Chiu, Wang, Song, Tibshirani, and Yeung-Levy]{lozano2025biomedica}
Alejandro Lozano, Min~Woo Sun, James Burgess, Liangyu Chen, Jeffrey~J. Nirschl, Jeffrey Gu, Ivan Lopez, Josiah Aklilu, Anita Rau, Austin~Wolfgang Katzer, Yuhui Zhang, Collin Chiu, Xiaohan Wang, Alfred~Seunghoon Song, Robert Tibshirani, and Serena Yeung-Levy.
\newblock {BIOMEDICA: An Open Biomedical Image-Caption Archive, Dataset, and Vision-Language Models Derived from Scientific Literature}.
\newblock In \emph{Proceedings of the IEEE/CVF Conference on Computer Vision and Pattern Recognition (CVPR)}, pages 19724--19735, 2025.

\bibitem[Lu et~al.(2022)Lu, Mishra, Xia, Qiu, Chang, Zhu, Tafjord, Clark, and Kalyan]{lu2022scienceqa}
Pan Lu, Swaroop Mishra, Tanglin Xia, Liang Qiu, Kai-Wei Chang, Song-Chun Zhu, Oyvind Tafjord, Peter Clark, and Ashwin Kalyan.
\newblock {Learn to Explain: Multimodal Reasoning via Thought Chains for Science Question Answering}.
\newblock In \emph{Advances in Neural Information Processing Systems}, pages 2507--2521, 2022.

\bibitem[Lu et~al.(2024)Lu, Bansal, Xia, Liu, Li, Hajishirzi, Cheng, Chang, Galley, and Gao]{lu2024mathvista}
Pan Lu, Hritik Bansal, Tony Xia, Jiacheng Liu, Chunyuan Li, Hannaneh Hajishirzi, Hao Cheng, Kai-Wei Chang, Michel Galley, and Jianfeng Gao.
\newblock {MathVista: Evaluating Mathematical Reasoning of Foundation Models in Visual Contexts}.
\newblock In \emph{International Conference on Learning Representations}, pages 23439--23554, 2024.

\bibitem[Ma et~al.(2026)Ma, Suo, Wang, and Zhang]{ma2026mmcothallucination}
Ji Ma, Wei Suo, Peng Wang, and Yanning Zhang.
\newblock {Understanding and Mitigating Hallucinations in Multimodal Chain-of-Thought Models}.
\newblock In \emph{Proceedings of the IEEE/CVF Conference on Computer Vision and Pattern Recognition (CVPR)}, pages 40224--40234, 2026.

\bibitem[Ma et~al.(2024)Ma, Zang, Chen, Chen, Jiao, Li, Lu, Liu, Ma, Dong, Zhang, Pan, Jiang, Wang, Cao, and Sun]{ma2024mmlongbench}
Yubo Ma, Yuhang Zang, Liangyu Chen, Meiqi Chen, Yizhu Jiao, Xinze Li, Xinyuan Lu, Ziyu Liu, Yan Ma, Xiaoyi Dong, Pan Zhang, Liangming Pan, Yu-Gang Jiang, Jiaqi Wang, Yixin Cao, and Aixin Sun.
\newblock {MMLONGBENCH-DOC: Benchmarking Long-context Document Understanding with Visualizations}.
\newblock In \emph{Advances in Neural Information Processing Systems}, pages 95963--96010, 2024.

\bibitem[Masry et~al.(2022)Masry, Long, Tan, Joty, and Hoque]{masry2022chartqa}
Ahmed Masry, Do~Xuan Long, Jia~Qing Tan, Shafiq Joty, and Enamul Hoque.
\newblock {{C}hart{QA}: A Benchmark for Question Answering about Charts with Visual and Logical Reasoning}.
\newblock In \emph{Findings of the {A}ssociation for {C}omputational {L}inguistics: {ACL} 2022}, pages 2263--2279, 2022.

\bibitem[Masry et~al.(2023)Masry, Kavehzadeh, Do, Hoque, and Joty]{masry2023unichart}
Ahmed Masry, Parsa Kavehzadeh, Xuan~Long Do, Enamul Hoque, and Shafiq Joty.
\newblock {{U}ni{C}hart: A Universal Vision-language Pretrained Model for Chart Comprehension and Reasoning}.
\newblock In \emph{Proceedings of the 2023 Conference on Empirical Methods in Natural Language Processing}, pages 14662--14684, 2023.

\bibitem[Masry et~al.(2024)Masry, Shahmohammadi, Parvez, Hoque, and Joty]{masry2024chartinstruct}
Ahmed Masry, Mehrad Shahmohammadi, Md~Rizwan Parvez, Enamul Hoque, and Shafiq Joty.
\newblock {{C}hart{I}nstruct: Instruction Tuning for Chart Comprehension and Reasoning}.
\newblock In \emph{Findings of the {A}ssociation for {C}omputational {L}inguistics: {ACL} 2024}, pages 10387--10409, 2024.

\bibitem[Masry et~al.(2025)Masry, Islam, Ahmed, Bajaj, Kabir, Kartha, Laskar, Rahman, Rahman, Shahmohammadi, Thakkar, Parvez, Hoque, and Joty]{masry2025chartqapro}
Ahmed Masry, Mohammed~Saidul Islam, Mahir Ahmed, Aayush Bajaj, Firoz Kabir, Aaryaman Kartha, Md~Tahmid~Rahman Laskar, Mizanur Rahman, Shadikur Rahman, Mehrad Shahmohammadi, Megh Thakkar, Md~Rizwan Parvez, Enamul Hoque, and Shafiq Joty.
\newblock {{C}hart{QAP}ro: A More Diverse and Challenging Benchmark for Chart Question Answering}.
\newblock In \emph{Findings of the {A}ssociation for {C}omputational {L}inguistics: {ACL} 2025}, pages 19123--19151, 2025.

\bibitem[Methani et~al.(2020)Methani, Ganguly, Khapra, and Kumar]{methani2020plotqa}
Nitesh Methani, Pritha Ganguly, Mitesh~M. Khapra, and Pratyush Kumar.
\newblock {PlotQA: Reasoning over Scientific Plots}.
\newblock In \emph{Proceedings of the IEEE/CVF Winter Conference on Applications of Computer Vision (WACV)}, pages 1527--1536, 2020.

\bibitem[Min et~al.(2026)Min, Zhou, Li, Wu, Peng, Du, Zhao, Yang, and Wen]{min2026perceptionprm}
Yingqian Min, Kun Zhou, Yifan Li, Yuhuan Wu, Han Peng, Yifan Du, Wayne~Xin Zhao, Min Yang, and Ji-Rong Wen.
\newblock {Improving Vision-language Models with Perception-centric Process Reward Models}.
\newblock In \emph{Proceedings of the IEEE/CVF Conference on Computer Vision and Pattern Recognition (CVPR)}, pages 33099--33109, 2026.

\bibitem[Pan et~al.(2026)Pan, Wu, Hong, Hu, Yan, Yang, Han, Zhou, Zhu, Han, and Chen]{pan2026chartfr1}
Hongkun Pan, Yuwei Wu, Wanyi Hong, Shenghui Hu, Qitong Yan, Yi Yang, Rufei Han, Changju Zhou, Minfeng Zhu, Dongming Han, and Wei Chen.
\newblock {Chart-FR1: Visual Focus-Driven Fine-Grained Reasoning on Dense Charts}.
\newblock In \emph{Proceedings of the IEEE/CVF Conference on Computer Vision and Pattern Recognition (CVPR)}, pages 26285--26294, 2026.

\bibitem[Panagopoulou et~al.(2025)Panagopoulou, Zhou, Savarese, Xiong, Callison-Burch, Yatskar, and Niebles]{panagopoulou2025viunit}
Artemis Panagopoulou, Honglu Zhou, Silvio Savarese, Caiming Xiong, Chris Callison-Burch, Mark Yatskar, and Juan~Carlos Niebles.
\newblock {ViUniT: Visual Unit Tests for More Robust Visual Programming}.
\newblock In \emph{Proceedings of the IEEE/CVF Conference on Computer Vision and Pattern Recognition (CVPR)}, pages 24646--24656, 2025.

\bibitem[Park et~al.(2025)Park, Kim, and Kim]{park2025halloc}
Eunkyu Park, Minyeong Kim, and Gunhee Kim.
\newblock {HalLoc: Token-level Localization of Hallucinations for Vision Language Models}.
\newblock In \emph{Proceedings of the IEEE/CVF Conference on Computer Vision and Pattern Recognition (CVPR)}, pages 29893--29903, 2025.

\bibitem[Pramanick et~al.(2024)Pramanick, Chellappa, and Venugopalan]{pramanick2024spiqa}
Shraman Pramanick, Rama Chellappa, and Subhashini Venugopalan.
\newblock {SPIQA: A Dataset for Multimodal Question Answering on Scientific Papers}.
\newblock In \emph{Advances in Neural Information Processing Systems}, pages 118807--118833, 2024.

\bibitem[Roberts et~al.(2024)Roberts, Han, Houlsby, and Albanie]{roberts2024scifibench}
Jonathan Roberts, Kai Han, Neil Houlsby, and Samuel Albanie.
\newblock {SciFIBench: Benchmarking Large Multimodal Models for Scientific Figure Interpretation}.
\newblock In \emph{Advances in Neural Information Processing Systems}, pages 18695--18728, 2024.

\bibitem[Shen et~al.(2025)Shen, Zhao, Zhao, Xu, Zhang, Zhu, and Yin]{shen2025zoomeye}
Haozhan Shen, Kangjia Zhao, Tiancheng Zhao, Ruochen Xu, Zilun Zhang, Mingwei Zhu, and Jianwei Yin.
\newblock {{Z}oom{E}ye: Enhancing Multimodal {LLM}s with Human-Like Zooming Capabilities through Tree-Based Image Exploration}.
\newblock In \emph{Proceedings of the 2025 Conference on Empirical Methods in Natural Language Processing}, pages 6602--6618, 2025.

\bibitem[Shen et~al.(2026)Shen, Qigqi, Ding, Xu, Meng, and Xiang]{shen2026scicqa}
Lingdong Shen, Qigqi, Kun Ding, Jian Xu, Gaofeng Meng, and Shiming Xiang.
\newblock {Advancing Scientific Chart Understanding: The SCI-CQA Benchmark and Beyond}.
\newblock \emph{IEEE Transactions on Image Processing}, 35:\penalty0 10829--10841, 2026.

\bibitem[Sur{\'\i}s et~al.(2023)Sur{\'\i}s, Menon, and Vondrick]{suris2023vipergpt}
D{\'\i}dac Sur{\'\i}s, Sachit Menon, and Carl Vondrick.
\newblock {ViperGPT: Visual Inference via Python Execution for Reasoning}.
\newblock In \emph{Proceedings of the IEEE/CVF International Conference on Computer Vision (ICCV)}, pages 11888--11898, 2023.

\bibitem[Tang et~al.(2025)Tang, Kim, Zhao, Lake, Ding, Yin, Singhal, Wadhwa, Liu, Sprague, Namuduri, Hu, Rodriguez, Peng, and Durrett]{tang2025chartmuseum}
Liyan Tang, Grace Kim, Xinyu Zhao, Thom Lake, Wenxuan Ding, Fangcong Yin, Prasann Singhal, Manya Wadhwa, Zeyu Liu, Zayne Sprague, Ramya Namuduri, Bodun Hu, Juan Rodriguez, Puyuan Peng, and Greg Durrett.
\newblock {{ChartMuseum}: Testing Visual Reasoning Capabilities of Large Vision-Language Models}.
\newblock In \emph{Advances in Neural Information Processing Systems}, 2025.

\bibitem[Tian et~al.(2025)Tian, Lu, Gao, Liu, and Zhao]{tian2025mmcr}
Yang Tian, Zheng Lu, Mingqi Gao, Zheng Liu, and Bo Zhao.
\newblock {{MMCR}: Benchmarking Cross-Source Reasoning in Scientific Papers}.
\newblock In \emph{Proceedings of the IEEE/CVF International Conference on Computer Vision (ICCV)}, pages 488--497, 2025.

\bibitem[Tu et~al.(2025)Tu, Hu, and Sang]{tu2025ode}
Yahan Tu, Rui Hu, and Jitao Sang.
\newblock {ODE: Open-Set Evaluation of Hallucinations in Multimodal Large Language Models}.
\newblock In \emph{Proceedings of the IEEE/CVF Conference on Computer Vision and Pattern Recognition (CVPR)}, pages 19836--19845, 2025.

\bibitem[Turpin et~al.(2023)Turpin, Michael, Perez, and Bowman]{turpin2023unfaithful}
Miles Turpin, Julian Michael, Ethan Perez, and Samuel Bowman.
\newblock {Language Models Don't Always Say What They Think: Unfaithful Explanations in Chain-of-Thought Prompting}.
\newblock In \emph{Advances in Neural Information Processing Systems}, pages 74952--74965, 2023.

\bibitem[Venna et~al.(2026)Venna, Gunda, Jinka, Rachakonda, Srinivasan, and Sarvadevabhatla]{venna2026m3grounder}
Venkata~Kesav Venna, Sai~Madhusudan Gunda, Jyothi~Swaroopa Jinka, Hrithik~Sagar Rachakonda, Anirudh Srinivasan, and Ravi~Kiran Sarvadevabhatla.
\newblock {M3Grounder: Mask-Based Multi-Span and Multi-Granular Grounding for Document QA}.
\newblock In \emph{Proceedings of the IEEE/CVF Conference on Computer Vision and Pattern Recognition (CVPR)}, pages 23685--23695, 2026.

\bibitem[Wang et~al.(2025{\natexlab{a}})Wang, Shen, Kuang, Cohan, and Zhao]{wang2025sciver}
Chengye Wang, Yifei Shen, Zexi Kuang, Arman Cohan, and Yilun Zhao.
\newblock {{S}ci{V}er: Evaluating Foundation Models for Multimodal Scientific Claim Verification}.
\newblock In \emph{Proceedings of the 63rd Annual Meeting of the {A}ssociation for {C}omputational {L}inguistics (Volume 1: Long Papers)}, pages 8562--8579, 2025{\natexlab{a}}.

\bibitem[Wang et~al.(2025{\natexlab{b}})Wang, Li, Yin, Ran, and Liu]{wang2025mvmath}
Peijie Wang, Zhong-Zhi Li, Fei Yin, Dekang Ran, and Cheng-Lin Liu.
\newblock {MV-MATH: Evaluating Multimodal Math Reasoning in Multi-Visual Contexts}.
\newblock In \emph{Proceedings of the IEEE/CVF Conference on Computer Vision and Pattern Recognition (CVPR)}, pages 19541--19551, 2025{\natexlab{b}}.

\bibitem[Wang et~al.(2024)Wang, Xia, He, Chen, Liu, Zhu, Liang, Wu, Liu, Malladi, Chevalier, Arora, and Chen]{wang2024charxiv}
Zirui Wang, Mengzhou Xia, Luxi He, Howard Chen, Yitao Liu, Richard Zhu, Kaiqu Liang, Xindi Wu, Haotian Liu, Sadhika Malladi, Alexis Chevalier, Sanjeev Arora, and Danqi Chen.
\newblock {CharXiv: Charting Gaps in Realistic Chart Understanding in Multimodal LLMs}.
\newblock In \emph{Advances in Neural Information Processing Systems}, pages 113569--113697, 2024.

\bibitem[Wang et~al.(2025{\natexlab{c}})Wang, Guan, Fu, Duan, Jiang, Guo, Guo, Luo, Shen, and Yang]{wang2025marten}
Zining Wang, Tongkun Guan, Pei Fu, Chen Duan, Qianyi Jiang, Zhentao Guo, Shan Guo, Junfeng Luo, Wei Shen, and Xiaokang Yang.
\newblock {Marten: Visual Question Answering with Mask Generation for Multi-modal Document Understanding}.
\newblock In \emph{Proceedings of the IEEE/CVF Conference on Computer Vision and Pattern Recognition (CVPR)}, pages 14460--14471, 2025{\natexlab{c}}.

\bibitem[Wei et~al.(2022)Wei, Wang, Schuurmans, Bosma, ichter, Xia, Chi, Le, and Zhou]{wei2022cot}
Jason Wei, Xuezhi Wang, Dale Schuurmans, Maarten Bosma, brian ichter, Fei Xia, Ed Chi, Quoc~V Le, and Denny Zhou.
\newblock {Chain-of-Thought Prompting Elicits Reasoning in Large Language Models}.
\newblock In \emph{Advances in Neural Information Processing Systems}, pages 24824--24837, 2022.

\bibitem[Wu and Xie(2024)]{wu2024vstar}
Penghao Wu and Saining Xie.
\newblock {V*: Guided Visual Search as a Core Mechanism in Multimodal LLMs}.
\newblock In \emph{Proceedings of the IEEE/CVF Conference on Computer Vision and Pattern Recognition (CVPR)}, pages 13084--13094, 2024.

\bibitem[Xia et~al.(2025)Xia, Ye, Yan, Liu, Zhou, Chen, Shi, Yan, and Zhang]{xia2025chartx}
Renqiu Xia, Hancheng Ye, Xiangchao Yan, Qi Liu, Hongbin Zhou, Zijun Chen, Botian Shi, Junchi Yan, and Bo Zhang.
\newblock {ChartX and ChartVLM: A Versatile Benchmark and Foundation Model for Complicated Chart Reasoning}.
\newblock \emph{IEEE Transactions on Image Processing}, 34:\penalty0 7436--7447, 2025.

\bibitem[Xiong et~al.(2026)Xiong, Ge, Li, Zhang, Kulkarni, Wang, He, Zhu, Liu, Chen, Zheng, Chen, Wang, Zhang, Chen, and Huang]{xiong2026multicrit}
Tianyi Xiong, Yi Ge, Ming Li, Zuolong Zhang, Pranav Kulkarni, Kaishen Wang, Qi He, Zeying Zhu, Chenxi Liu, Ruibo Chen, Tong Zheng, Yanshuo Chen, Xiyao Wang, Renrui Zhang, Wenhu Chen, and Heng Huang.
\newblock {Multi-Crit: Benchmarking Multimodal Judges on Pluralistic Criteria-Following}.
\newblock In \emph{Proceedings of the IEEE/CVF Conference on Computer Vision and Pattern Recognition (CVPR)}, pages 8641--8652, 2026.

\bibitem[Xu et~al.(2025{\natexlab{a}})Xu, Zhu, and Yang]{xu2025mcbench}
Yunqiu Xu, Linchao Zhu, and Yi Yang.
\newblock {{MC-Bench}: A Benchmark for Multi-Context Visual Grounding in the Era of {MLLMs}}.
\newblock In \emph{Proceedings of the IEEE/CVF International Conference on Computer Vision (ICCV)}, pages 17675--17687, 2025{\natexlab{a}}.

\bibitem[Xu et~al.(2025{\natexlab{b}})Xu, Du, Qi, Lu, Xu, Yuan, and Guo]{xu2025chartpoint}
Zhengzhuo Xu, SiNan Du, Yiyan Qi, Siwen Lu, Chengjin Xu, Chun Yuan, and Jian Guo.
\newblock {ChartPoint: Guiding MLLMs with Grounding Reflection for Chart Reasoning}.
\newblock In \emph{Proceedings of the IEEE/CVF International Conference on Computer Vision (ICCV)}, pages 426--436, 2025{\natexlab{b}}.

\bibitem[Yang et~al.(2025)Yang, Zhang, Hou, Li, Liu, Payani, Ting, and Zheng]{yang2025chartsynthesis}
Yuwei Yang, Zeyu Zhang, Yunzhong Hou, Zhuowan Li, Gaowen Liu, Ali Payani, Yuan-Sen Ting, and Liang Zheng.
\newblock {Effective Training Data Synthesis for Improving MLLM Chart Understanding}.
\newblock In \emph{Proceedings of the IEEE/CVF International Conference on Computer Vision (ICCV)}, pages 2653--2663, 2025.

\bibitem[Yuan et~al.(2026)Yuan, Sun, Chen, Lozano, He, Li, Navab, Sun, Padoy, and Yeung-Levy]{yuan2026panelpixel}
Kun Yuan, Min Sun, Zhen Chen, Alejandro Lozano, Xiangteng He, Shi Li, Nassir Navab, Xiaoxiao Sun, Nicolas Padoy, and Serena Yeung-Levy.
\newblock {From Panel to Pixel: Zoom-In Vision-Language Pretraining from Biomedical Scientific Literature}.
\newblock In \emph{Proceedings of the IEEE/CVF Conference on Computer Vision and Pattern Recognition (CVPR)}, pages 42649--42658, 2026.

\bibitem[Yue et~al.(2024)Yue, Ni, Zhang, Zheng, Liu, Zhang, Stevens, Jiang, Ren, Sun, Wei, Yu, Yuan, Sun, Yin, Zheng, Yang, Liu, Huang, Sun, Su, and Chen]{yue2024mmmu}
Xiang Yue, Yuansheng Ni, Kai Zhang, Tianyu Zheng, Ruoqi Liu, Ge Zhang, Samuel Stevens, Dongfu Jiang, Weiming Ren, Yuxuan Sun, Cong Wei, Botao Yu, Ruibin Yuan, Renliang Sun, Ming Yin, Boyuan Zheng, Zhenzhu Yang, Yibo Liu, Wenhao Huang, Huan Sun, Yu Su, and Wenhu Chen.
\newblock {MMMU: A Massive Multi-discipline Multimodal Understanding and Reasoning Benchmark for Expert AGI}.
\newblock In \emph{Proceedings of the IEEE/CVF Conference on Computer Vision and Pattern Recognition (CVPR)}, pages 9556--9567, 2024.

\bibitem[Zheng et~al.(2023)Zheng, Chiang, Sheng, Zhuang, Wu, Zhuang, Lin, Li, Li, Xing, Zhang, Gonzalez, and Stoica]{zheng2023mtbench}
Lianmin Zheng, Wei-Lin Chiang, Ying Sheng, Siyuan Zhuang, Zhanghao Wu, Yonghao Zhuang, Zi Lin, Zhuohan Li, Dacheng Li, Eric Xing, Hao Zhang, Joseph Gonzalez, and Ion Stoica.
\newblock {Judging LLM-as-a-Judge with MT-Bench and Chatbot Arena}.
\newblock In \emph{Advances in Neural Information Processing Systems}, pages 46595--46623, 2023.

\end{thebibliography}
}
\end{document}